# Automatic Skin Lesion Analysis using Large-scale Dermoscopy Images and Deep Residual Networks

Lei Bi, Jinman Kim, Euijoon Ahn, and Dagan Feng

BMIT Research Lab, School of Information Technologies, the University of Sydney, Australia
firstname.lastname@sydney.edu.au

## 1 Introduction

Malignant melanoma has one of the most rapidly increasing incidences in the world and has a considerable mortality rate. Early diagnosis is particularly important since melanoma can be cured with prompt excision. Dermoscopy images play an important role in the non-invasive early detection of melanoma [1]. However, melanoma detection using human vision alone can be subjective, inaccurate and poorly reproducible even among experienced dermatologists. This is attributed to the challenges in interpreting images with diverse characteristics including lesions of varying sizes and shapes, lesions that may have fuzzy boundaries, different skin colors and the presence of hair [2]. Therefore, the automatic analysis of dermoscopy images is a valuable aid for clinical decision making and for image-based diagnosis to identify diseases such as melanoma [1-4].

## 2 Methods

### 2.1 Deep Residual Networks

Deep residual networks (ResNets) has achieved state-of-the-art results in image classification and detection related problems [5-8]. Compared with many deep networks, e.g., VGGNet, adding extra layers (beyond certain depth) results in higher training and validation errors. Therefore, it is challenging to optimize a very deep networks with many layers. ResNets architecture consists of a number of residual blocks with each block comprising of several convolution layers, batch normalization layers and ReLU layers. The residual block enables to bypass (shortcut) a few convolution layers at a time [5]. Therefore, the ResNets architecture is capable to overcome this limitation by adding shortcut connections that are aggregated with the output of the convolution layers [5]. In addition, ResNets can be considered as an ensembles of many networks [6-8], where different networks are connected via these shortcuts and therefore, optimal results can be achieved by averaging the output of the different networks. In this ISIC 2017 skin lesion analysis challenge [9], we propose to exploit the deep ResNets for robust visual features learning and representations.

## 2.2 Lesion Segmentation

Lesion segmentation is a fundamental requirement for automated skin lesion analysis [10, 11]. For segmentation, we followed the fully convolutional networks (FCN) architecture [12] to add convolutional and deconvolutional layers, which upsample the features maps derived from ResNet to output the score mask.

Besides the training images (2,000 training images) provided by the challenge organizer, we acquired additional ~8,000 images from the international skin imaging collaboration (ISIC) archive [13]. We only used the images annotated by experts (indicated by the image attribute file). In total, we used ~9,800 images for training and ~200 images for training validation (tuning the parameters). All the images were downscaled to 500 pixels (longer axis). During training, we first fine-tuned the ResNet model, pre-trained on ImageNet dataset [14] for ~60 epochs using a fixed learning rate of 0.0016. After that, we further fine-tuned the model using the 2017 challenge training images (2,000 images) for another ~80 epochs with a linear schedule learning rate at base of 0.0008. Data argument including random crops and flips were used to improve the robustness of the model [15, 16]. The training image batch size was set to 10 and the whole training process took around 7 days on two Titan X GPUs. At the testing time, a multi-scale integration approach was used, where we resize the image into a number of scales. On each scale, we argument the input image by flip left-right, upside-down and both flip left-right and upside-down. The final output was produced by integrating the multi-scale outputs.

## 2.3 Lesion Classification

The aim of the classification is to label skin lesions as melanoma, seborrheic keratosis and nevus. Melanoma is defined as malignant skin tumor, derived from melanocytes (melanocytic); Seborrheic keratosis (SK) is defined as benign skin tumor, derived from keratinocytes (non-melanocytic); and Nevus is defined as benign skin tumor, derived from melanocytes (melanocytic).

We developed three separated approaches. In the first approach, we considered the classification as a multi-class (3 labels) classification problem. In the second approach, we considered the classification as two binary classification problems, where the two binary classifiers were separating melanoma from the others and separating seborrheic keratosis from the others. In the third approach, we ensemble the first two approaches to produce the final results.

Similar to the skin lesion segmentation task, we acquired another ~1,600 images (histology or expert confirmed studies) from the ISIC archive [13], which resulted in a total of ~3,600 images for training (2,000 training images were provided by the challenge organizer). We modified the last layer of ResNet to have 3 or 2 neurons to match the number of classes. All the images were downscaled to 224 pixels (shorter axis). During training, we fine-tuned the ResNet model, pre-trained on ImageNet dataset for

150 epochs using a linear schedule learning rate at a base learning rate of 0.01. Data argument including random crops and flips were used to improve the robustness of the model [15, 16]. The training image batch size was set to 90. The training of one ResNet took around half day on two Titan X GPUs.

## 3  Experiments and Results

### 3.1  Experimental Setup

The evaluation was conducted on the test validation set, provided by the challenge organizer. In total, there are 150 skin lesion images; the ground truth are not available to the public. The results were processed by the online submission system.

### 3.2  Lesion Segmentation Results

We compared the proposed segmentation method (ResNet-Seg) with the FCN architecture (VGGNet model). In addition, we evaluated the performance of using the additional training images and also the proposed multi-scale integration approach.

The proposed multi-scale approach achieved the 1st place on the validation set by the submission deadline[1].

**Table 1.** Comparison of the skin lesion segmentation for different methods.

| % | With additional Training Images | Jaccard |
|---|---|---|
| **FCN** | No | 68.00 |
| **ResNet-Seg** | No | 76.10 |
| **ResNet-Seg** | Yes | 77.80 |
| **MResNet-Seg (Multi-scale)** | Yes | **79.40** |

### 3.3  Lesion Classification Results

We evaluated the proposed three classification approaches (1) multi-class classification; (2) binary classification approach; and (3) the ensemble approach, based on the 150 validation images. Averaged area under curve (AUC) was used for evaluation.

**Table 2.** Comparison of the skin lesion classification for different methods.

| % | With additional Training Images | Melanoma AUC | SK AUC | Average AUC |
|---|---|---|---|---|
| **ResNet (multi-class)** | Yes | 84.30 | 96.90 | 90.60 |
| **ResNet (binary)** | Yes | **85.50** | 97.00 | 91.30 |
| **ResNet (Ensembled)** | Yes | 85.40 | **97.60** | **91.50** |

---

[1.] https://challenge.kitware.com/#phase/584b08eecad3a51cc66c8e1f